\documentclass{article}

\PassOptionsToPackage{numbers, compress}{natbib}

\usepackage[preprint]{neurips_2019}

\usepackage[utf8]{inputenc} 
\usepackage[T1]{fontenc}    
\usepackage{hyperref}       
\usepackage{url}            
\usepackage{booktabs}       
\usepackage{amsfonts}       
\usepackage{nicefrac}       
\usepackage{microtype}      

\usepackage{color}
\usepackage{enumitem}
\usepackage{graphicx}
\usepackage{arydshln}
\usepackage{multirow}
\usepackage{bm}
\usepackage{epsfig}
\usepackage{amsmath}
\usepackage{amssymb}
\usepackage{algorithm}
\usepackage{algorithmicx}
\usepackage{algpseudocode}
\usepackage{caption}
\usepackage{subcaption}
\usepackage{graphicx}
\usepackage{float}
\usepackage{wrapfig}
\usepackage{authblk}
\makeatletter
\renewcommand\AB@affilsepx{ \ \ \ \ \ \ \ \ \  \protect\Affilfont}
\makeatother

\title{Spatial Group-wise Enhance: Improving Semantic Feature Learning in Convolutional Networks}

\author[1,2]{Xiang Li\thanks{Xiang Li and Jian Yang are with PCA Lab, Key Lab of Intelligent Perception and Systems for High-Dimensional Information of Ministry of Education, and Jiangsu Key Lab of Image and Video Understanding for Social Security, School of Computer Science and Engineering, Nanjing University of Science and Technology, China. Xiang Li is also a visiting scholar at Momenta. Email: xiang.li.implus@njust.edu.cn}\ \ }
\author[3]{Xiaolin Hu\thanks{Xiaolin Hu is with the Tsinghua National Laboratory for Information Science and Technology (TNList) Department of Computer Science and Technology, Tsinghua University, China.}\ \ }
\author[1]{Jian Yang\thanks{Corresponding author.}\ \ }
\affil[1]{PCALab, Nanjing University of Science and Technology}
\affil[2]{Momenta}
\affil[3]{Tsinghua University}

\begin{document}
	
	\maketitle
	
	\begin{abstract}
		The Convolutional Neural Networks (CNNs) generate the feature representation of complex objects by collecting hierarchical and different parts of semantic sub-features. These sub-features can usually be distributed in grouped form in the feature vector of each layer \cite{wu2018group,sabour2017dynamic}, representing various semantic entities. However, the activation of these sub-features is often spatially affected by similar patterns and noisy backgrounds, resulting in erroneous localization and identification. We propose a Spatial Group-wise Enhance (SGE) module that can adjust the importance of each sub-feature by generating an attention factor for each spatial location in each semantic group, so that every individual group can autonomously enhance its learnt expression and suppress possible noise. The attention factors are only guided by the similarities between the global and local feature descriptors inside each group, thus the design of SGE module is extremely lightweight with \emph{almost no extra parameters and calculations}. Despite being trained with only category supervisions, the SGE component is extremely effective in highlighting multiple active areas with various high-order semantics (such as the dog's eyes, nose, etc.). When integrated with popular CNN backbones, SGE can significantly boost the performance of image recognition tasks. Specifically, based on ResNet50 backbones, SGE achieves 1.2\% Top-1 accuracy improvement on the ImageNet benchmark and 1.0$\sim$2.0\% AP gain on the COCO benchmark across a wide range of detectors (Faster/Mask/Cascade RCNN and RetinaNet). Codes and pretrained models are available at https://github.com/implus/PytorchInsight.
	\end{abstract}
	
	\section{Introduction}
	The idea of grouping features is long-standing. In the early research of computer vision, many artificially designed image features are presented in groups, such as SIFT \cite{lowe2004distinctive}, HOG \cite{dalal2005histograms}. For example, a HOG vector comes from several spatial cells where each cell is represented by a normalized orientation histogram. With the rapid development of CNNs \cite{lecun1990handwritten,krizhevsky2012imagenet,simonyan2014very,szegedy2015going,he2016deep,huang2017densely,wang2018mixed}, there are widely used module designs that introduce the grouping methodology, such as group convolution \cite{xie2017aggregated} and group normalization \cite{wu2018group}. These techniques typically group features along the channel dimension in a convolutional feature map into multiple sub-features, and use general convolution or normalization for the transformations of these sub-features in each group. In CapsuleNet \cite{sabour2017dynamic}, the grouped sub-features are modeled as capsules, which represent the instantiation parameters of a specific type of entity, such as an object or an object part.
	
\begin{figure}[t]
		\begin{center}
			\setlength{\fboxrule}{0pt}
			\fbox{\includegraphics[width=\textwidth]{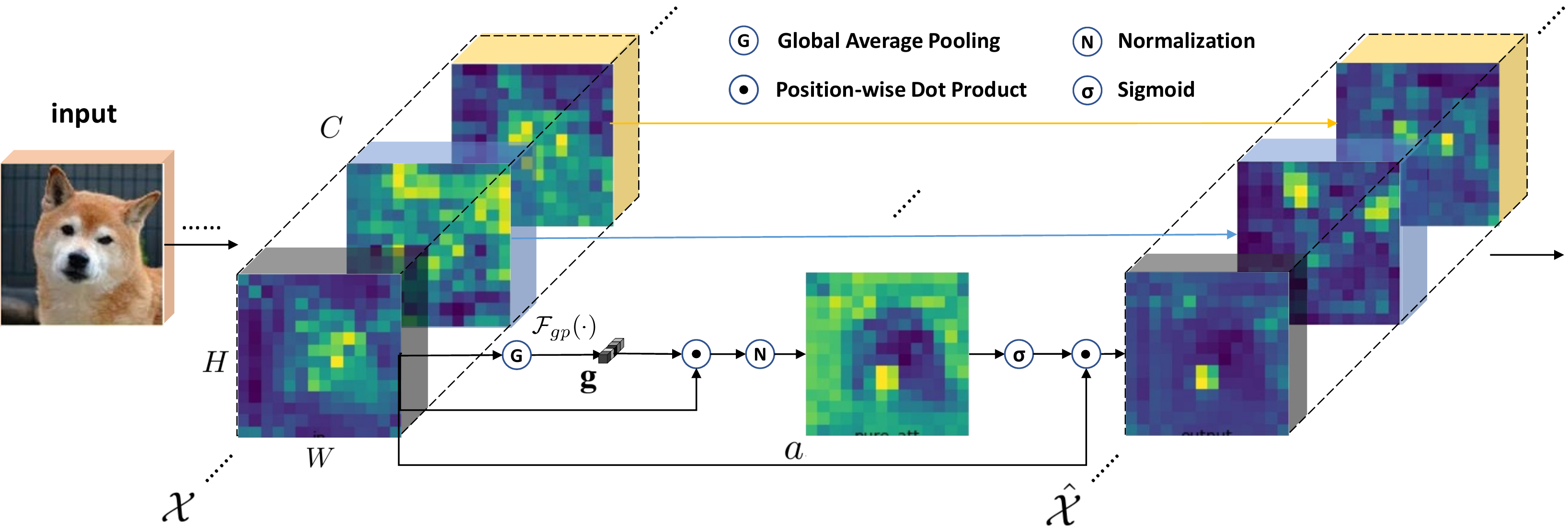}}
		\end{center}	
		\vspace{-8pt}
		\caption{Illustration of the proposed lightweight SGE module. It processes the sub-features of each group in parallel, and uses the similarity between global statistical feature and local positional features in each group as the attention guidance to enhance the features, thus obtaining well-distributed semantic feature representations in space.}
		\label{fig_sge}
		\vspace{-8pt}
	\end{figure}

	In addition to grouping the dimension of channels into multiple sub-features to represent different semantics, we also need to consider another important dimension in the convolutional feature map: the space. For a particular semantic group, it is reasonable and beneficial to generate the corresponding semantic features in the correct spatial positions of the original image. However, due to lack of supervision of specific region details and possible noise in the image, the spatial distribution of the semantic features will suffer from certain chaos, which considerably weakens the representation of learning and makes it difficult in constructions of hierarchical understanding (see $\mathcal{X}$ of Figure \ref{fig_sge}).
	
	In order to make each set of features robust and well-distributed over the space, we model a spatial enhance mechanism inside each feature group, by scaling the feature vectors over all the locations with an attention mask. Such an attention mask is designed intentionally to suppress the possible noise and highlight the correct semantic feature regions. Different from other popular attention methods \cite{wang2017residual,hu2018squeeze,li2019selective,park2018bam,woo2018cbam}, we use the similarity between the global statistical feature and the local ones of each location as the source of generation for the attention masks. This simple yet effective mechanism described above is our proposed Spatial Group-wise Enhance (SGE) module, which is extremely lightweight and requires \emph{almost no additional parameters and calculations} by its nature.
	
	We examine the changes in the distribution of the feature map and the statistics of the variance of the activation values for each group after the introduction of the SGE module. The results show that SGE significantly improves the spatial distribution of different semantic sub-features within its group, and produces a large variance statistically, which strengthens the feature learning in semantic regions and compresses the noise and interference.
	
	We show on the ImageNet \cite{russakovsky2015imagenet} benchmark that the SGE module performs better or comparable to a series of recently proposed state-of-the-art attention modules, despite its superiority in both model capacity and complexity. Meanwhile, for the most advanced detectors (Faster/Mask/Cascade RCNN \cite{ren2015faster,he2017mask,cai2018cascade}), SGE can always bring more than 1\% AP gains on the COCO \cite{lin2014microsoft} benchmark. Notably, on RetinaNet \cite{lin2017focal}, SGE outperforms the widely used SE \cite{hu2018squeeze} module on detecting small objects by $\sim$1\% AP, which demonstrates its remarkable advantages in accurate spatial modeling.
	\section{Related Work}
	\textbf{Grouped Features.} Learning and distributing features into groups in convolutional networks has been widely studied recently. AlexNet \cite{krizhevsky2012imagenet} initially presents the {group convolution} and divides features into two groups on different GPUs to save computing budgets. ResNeXt \cite{xie2017aggregated} examines the importance of grouping in feature transfer and suggests that the number of groups should be increased to obtain higher accuracy under similar model complexity. The MobileNet series \cite{howard2017mobilenets,sandler2018mobilenetv2,howard2019searching} and Xception \cite{carreira1998xception} treat each channel as a group and model only spatial relationships inside these groups. The ShuffleNet \cite{zhang1707shufflenet,ma2018shufflenet} family rearranges the grouped features to produce efficient feature representation. Res2Net \cite{gao2019res2net} uses a hierarchical mode to transfer grouped sub-features, enabling the network to incorporate multi-scale features in a single bottleneck. CapsuleNet \cite{sabour2017dynamic} models each of the grouped neurons as a capsule, where the activities of the neurons within an active capsule represent the various properties of a particular entity that is present in the image. The overall length of the vector of instantiation parameters is used to represent the existence of the entity and the orientation of the vector is forced to represent the properties of the entity. In SGE, all enhancements are operated inside groups, which saves computational overhead similarly as in {group convolution}. Conceptually, the SGE module adopts the basic modeling assumptions of CapsuleNet, and believes that the features of each group are able to actively learn various semantic entity representations. At the same time, in the process of visualization of this paper, we also use the length of the sub-feature to measure as its activation value, analogous to the probability of the existence of entities in CapsuleNet.
	
\begin{figure}[t]
		\begin{center}
			\setlength{\fboxrule}{0pt}
			\fbox{\includegraphics[width=\textwidth]{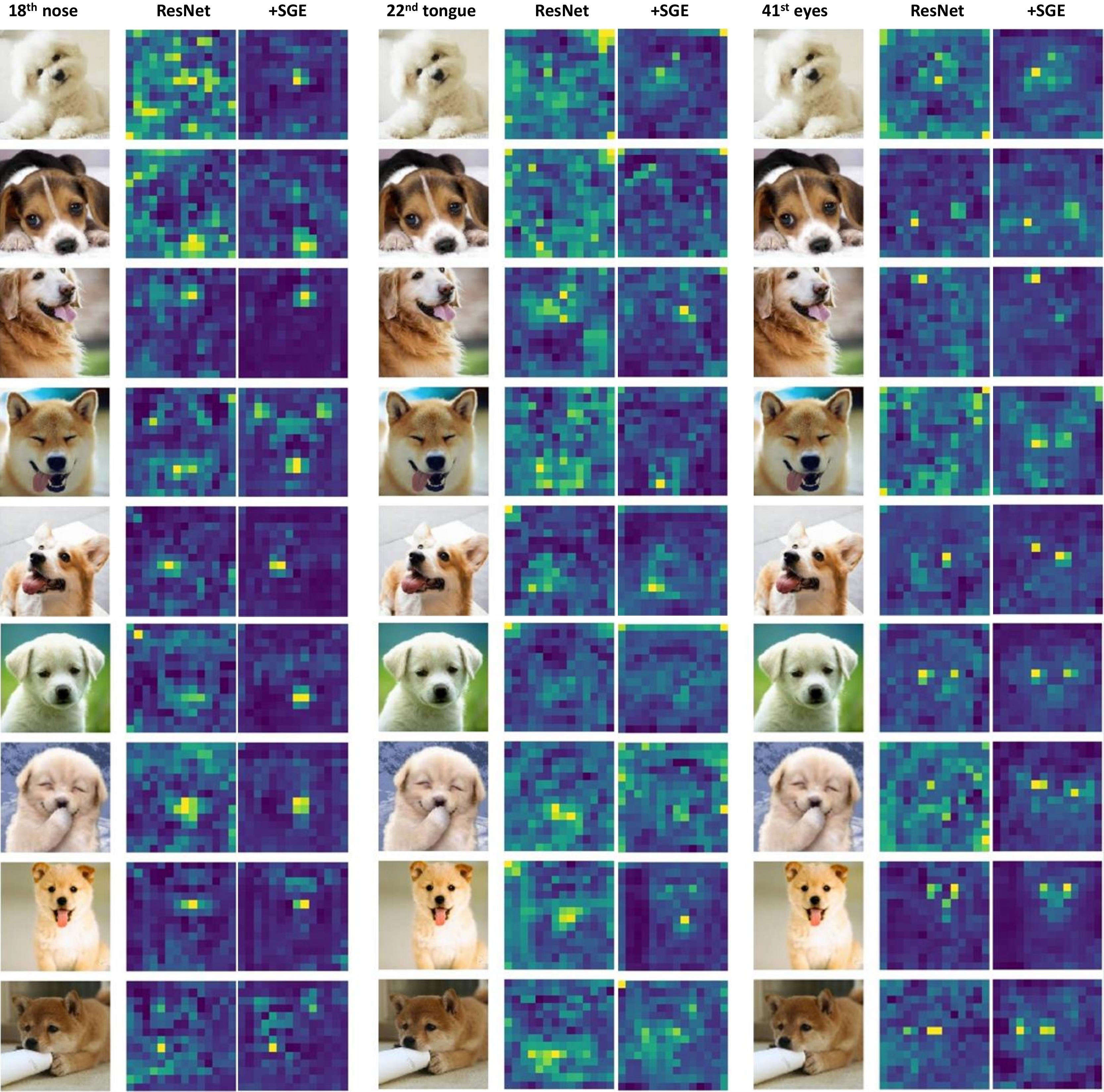}}
		\end{center}	
		\vspace{-10pt}
		\caption{We select several feature groups with representative semantics to display before and after using SGE on ResNet50. The semantics of the activated regions are found to be the nose from the 18th group, the tongue from the 22nd group, and the eyes from the 41st group, respectively. We sample images of different shapes, categories, and angles to verify the robustness of the SGE module.}
		\label{fig_vis}
		\vspace{-10pt}
	\end{figure}

	\textbf{Attention Models.} Attention models have recently become very popular. It first attracts widespread attention from the field of machine translation \cite{bahdanau2014neural,vaswani2017attention} and is later extended to more natural language processing tasks such as text summary \cite{rush2015neural} and reading comprehension \cite{seo2016bidirectional}. Since then, it has also achieved very promising results in the field of computer vision with emerging applications, such as person re-ID \cite{chen2018person}, image recovery \cite{zhang2018image}, lip reading \cite{xu2018lcanet}, image classification \cite{wang2017residual}, and object detection \cite{cao2019GCNet}. SENet \cite{hu2018squeeze} brings an effective, lightweight gating mechanism to self-recalibrate the feature map via channel-wise importances. Beyond channel, BAM \cite{park2018bam} and CBAM \cite{woo2018cbam} introduce spatial attention in a similar way. SKNet \cite{li2019selective} further introduces a dynamic kernel selection mechanism which is guided by the multi-scale group convolutions, with a small number of additional parameters and calculations to improve the classification performance. GCNet \cite{cao2019GCNet} fully explores the advantages and disadvantages of Non-Local \cite{wang2018non} and SE \cite{hu2018squeeze} modules, and combines the advantages of both to design a more effective global context module, obtaining compelling results on object detection tasks. SGE differs from all existing attention mechanisms in that it aims at improving the learning of different semantic sub-features of each group, intentionally self-enhancing its spatial distribution within the group. Compared to other attention modules, SGE has fewer parameters, less computational complexity (Table \ref{table_imagenet_cls}), and a more interpretable mechanism (Figure \ref{fig_vis}).
	
	\section{Method}
	\subsection{Spatial Group-wise Enhance}
	We consider a $C$ channels, $H \times W$ convolutional feature map and divide it into $G$ groups along the channel dimension. Without loss of generality, we first examine a certain group separately (see the bottom black box in Figure \ref{fig_sge}). Then the group has a vector representation at every position in space, namely  $\mathcal{X} = \left\{\textbf{x}_{1 \ldots m}\right\}, \textbf{x}_i \in \mathbb{R}^{\frac{C}{G}}, m = H \times W$. Conceptually inspired by the capsules \cite{sabour2017dynamic}, we further assume that this group gradually captures a specific semantic response (such as the dog's eyes) during the course of network learning. In this group space, ideally we can get features with strong responses at the eye positions (i.e., features with a larger vector length and similar vector directions among multiple eye regions), whilst other positions almost have no activation and become zero vectors. However, due to the unavoidable noise and the existence of similar patterns, it is usually difficult for CNNs to obtain the well-distributed feature responses. To address this issue, we propose to utilize the overall information of the entire group space to further enhance the learning of semantic features in critical regions, given the fact that the features of the entire space are not dominated by noise (otherwise the model learns nothing from this group). Therefore we can use the global statistical feature through spatial averaging function $\mathcal{F}_{gp}(\cdot)$ to approximate the semantic vector that this group learns to represent:
	\begin{equation}
	\textbf{g} = \mathcal{F}_{gp}(\mathcal{X}) = \frac{1}{m}\sum_{i=1}^{m}{\textbf{x}_{i}}.
	\label{eq_u}
	\end{equation}
	Next, using this global feature, we can generate the corresponding importance coefficient for each feature, which is obtained by simple dot product that measures the similarity between the global semantic feature $\textbf{g}$ and local feature $\textbf{x}_{i}$ to some extent. Thereby for each position, we have:
	\begin{equation}
	c_{i} = \textbf{g} \cdot \textbf{x}_{i}.
	\end{equation}
	Note that $c_i$ can also be expanded as $\|\textbf{g}\| \|\textbf{x}_{i}\| \cos(\theta_i)$, where $\theta_i$ is the angle between $\textbf{g}$ and $\textbf{x}_{i}$. It indicates that features that have a larger vector length (i.e., $\|\textbf{x}_{i}\|$) and a direction (i.e., $\theta_i$) closer to $\textbf{g}$ are more likely to obtain a larger initial coefficient, which is in line with our assumptions.  In order to prevent the biased magnitude of coefficients between various samples, we normalize $c$ over the space, as is widely practiced in \cite{ioffe2015batch,wu2018group,weightstandardization}:
	\begin{equation}
	\hat{c}_i=\frac{c_i -\mu_{c}}{\sigma_{c}+\epsilon},\ \  \mu_{c}=\frac{1}{m} \sum_{j}^{m} c_{j}, \ \   \sigma_{c}^2 =\frac{1}{m} \sum_{j}^{m} (c_j - \mu_{c})^2,
	\end{equation}
	where $\epsilon$ (e.g., 1e-5) is a constant added for numerical stability. To make sure that the normalization inserted in the network can represent the identity transform, we introduce a pair of parameters $\gamma, \beta$ for each coefficient $\hat{c}_i$, which scale and shift the normalized value:
	\begin{equation}
	{a}_i=\gamma \hat{c}_i +\beta.
	\end{equation}
	Note that $\gamma, \beta$ here are the only parameters introduced in our module. In a single SGE unit, the number of $\gamma, \beta$ is the same as the number of groups $G$, and the order of their magnitude is about tens (typically, 32 or 64), which is basically \emph{negligible} compared to the millions of parameters of the entire network. Finally, to obtain the enhanced feature vector $\hat{\textbf{x}}_{i}$, the original $\textbf{x}_{i}$ is scaled by the generated importance coefficients ${a}_i$ via a sigmoid function gate $\mathcal{\sigma}(\cdot)$ over the space:
	\begin{equation}
	\hat{\textbf{x}}_{i} = \textbf{x}_{i} \cdot \mathcal{\sigma}({a}_i),
	\end{equation}
	and all the enhanced features form the resulted feature group $\hat{\mathcal{X}} = \left\{\hat{\textbf{x}}_{1 \ldots m}\right\}, \hat{\textbf{x}}_i \in \mathbb{R}^{\frac{C}{G}}, m = H \times W$.

	\subsection{Visualization and Interpretation}
	\textbf{Visualization of Semantic Activation. }
	In order to verify whether our approach achieves the goal of improving the semantic feature representation, we train a network based on ResNet50 on ImageNet \cite{russakovsky2015imagenet} and place the SGE module after the last BatchNorm \cite{ioffe2015batch} layer of each bottleneck with reference to SENet \cite{hu2018squeeze}, by setting $G$ = 64. To better reflect the semantic information while preserving the large spatial resolution as much as possible, we choose to examine the feature maps of the 4th stage with output size of 14 $\times$ 14. For each feature vector of each group, we use its length (i.e., $\|\textbf{x}_{i}\|$) to indicate their activation value and linearly normalize it to the interval $[0, 1]$ for a better view. Figure \ref{fig_vis} shows three representative groups with semantic responses. As listed in three large columns, they are the 18th, 22nd, and 41st group, which are empirically found to correspond to the concept of the nose, tongue, and eyes. Each large column contains three small columns, where the first small column is the original image, the second small column is the feature map response from the original ResNet50, and the third one is the feature map response enhanced by the SGE module. We select images of dogs of different angles and types to test the robustness of SGE for feature enhancement. Despite its simplicity, the SGE module is very effective in improving the feature representation of specific semantics at corresponding locations while suppressing a large amount of noise. It is worth noting that in the 4th and 7th rows, SGE can strongly emphasize the activation of the eye areas, although their eyes are almost closed. In contrast, the original ResNet fails to capture such patterns. 
	
	\begin{wrapfigure}{l}{0.36\textwidth}
		\vspace{-16pt}
		\begin{center}
			\includegraphics[width=0.36\textwidth]{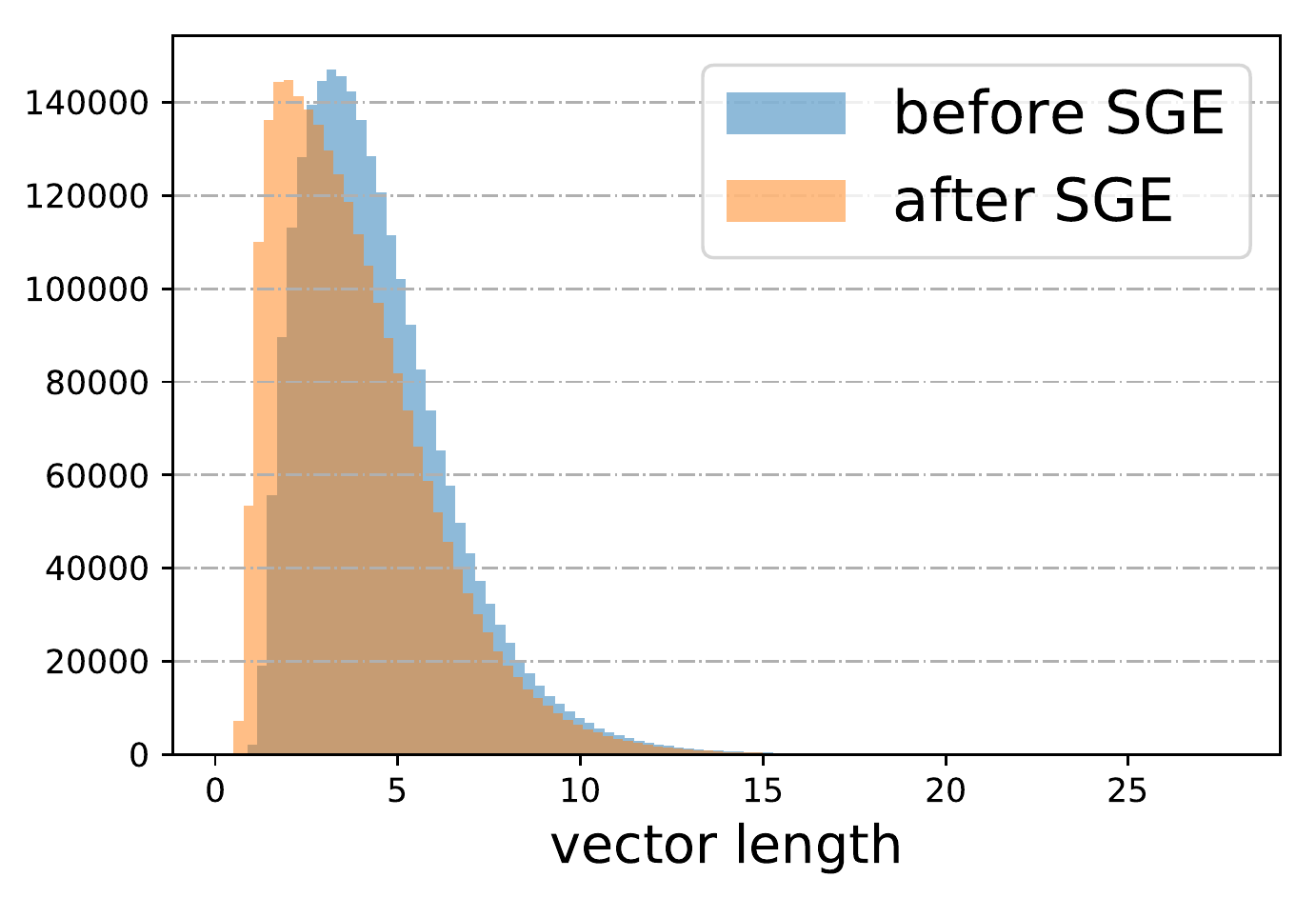}
		\end{center}
		\vspace{-12pt}	
		\caption{Histogram of activations.}
		\label{fig_hist}
		\vspace{-12pt}
	\end{wrapfigure}
	
\begin{figure}[t]
		\begin{center}
			\setlength{\fboxrule}{0pt}
			\fbox{\includegraphics[width=\textwidth]{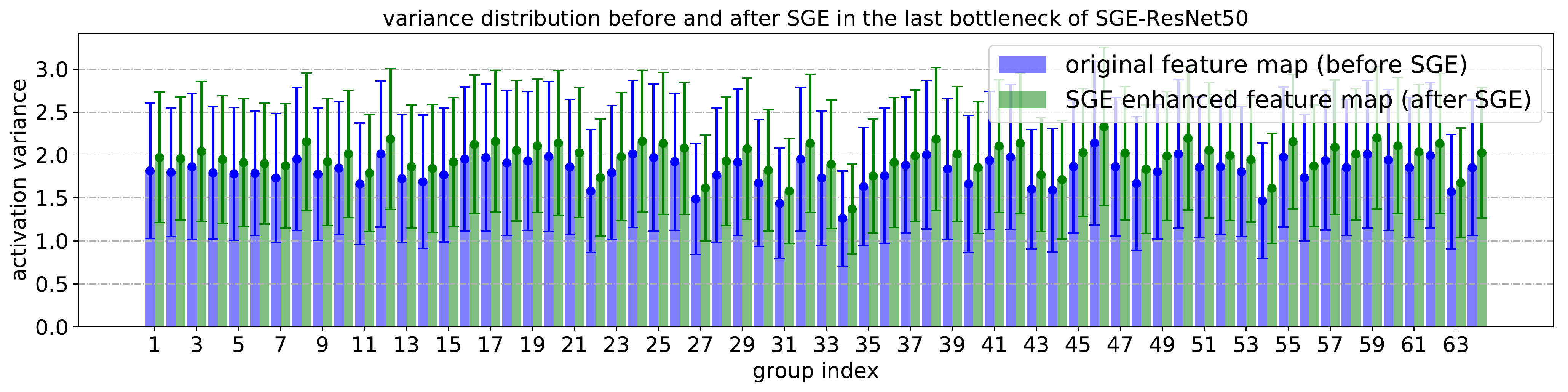}}
		\end{center}	
		\vspace{-14pt}
		\caption{The distribution of variance of activation values of each group, from the feature maps before and after SGE module in the last bottleneck of SGE-ResNet50. Standard deviation is also plotted.}
		\label{fig_vd}
		\vspace{-12pt}
	\end{figure}

	\textbf{The Statistical Change of Activation.} We note that if the ideal feature map is obtained, the spatial activation value of the network will have a more pronounced contrast, such as a large or sharp numerical activation in the semantically relevant regions, and nearly no response in other non-correlated regions. This contrast may probably correspond to a large degree of variance or sparsity to some extent. To validate this, we take the length of each sub-feature (i.e., $\|\textbf{x}_{i}\|$) as the activation value, and calculate their distribution of variance in each group of the last (highest) residual layer before and after using the SGE module. These statistics are based on the pretrained SGE-ResNet50, using all the samples on ImageNet validation set (i.e., 50k samples). As shown in Figure \ref{fig_vd}, the statistical results are in line with our expectations. The response variance of the feature map enhanced by the SGE module is indeed statistically increased, which greatly improves the efficiency of SGE to accurately capture semantic features. Furthermore, we plot the detailed histogram of the activation values of the first group over each position and all validation samples in Figure \ref{fig_hist}. It is observed that the smaller activation values bias towards zero and larger activation values nearly remain unchanged, which statistically implies the noise suppression and critical-region enhancement.

	\section{Experiments on Image Classification}
	We first compare SGE with a set of state-of-the-art attention modules on ImageNet benchmark. The ImageNet 2012 dataset \cite{russakovsky2015imagenet} comprises 1.28 million training images and 50k validation images from 1k classes. We train networks on the training set and report the Top-1 and Top-5 accuracies on the validation set with single 224 $\times$ 224 central crop. For data augmentation, we follow the standard practice \cite{szegedy2015going} and perform the random-size cropping to 224 $\times$ 224 and random horizontal flipping. The practical mean channel subtraction is adopted to normalize the input images. All networks are trained with naive softmax cross entropy without label-smoothing regularization \cite{szegedy2016rethinking}. We train all the architectures from scratch by synchronous SGD with weight decay 0.0001 and momentum 0.9 for 100 epochs, starting from learning rate 0.1 and decreasing it by a factor of 10 every 30 epochs. The total batch size is set as 256 and 8 GPUs (32 images per GPU) are utilized for training, using the weight initialization strategy in \cite{he2015delving}. Our codes are implemented in the pytorch \cite{paszke2017pytorch} framework. Note that in the following tables, Param. denotes the number of parameter and the definition of FLOPs follow  \cite{zhang1707shufflenet}, i.e., the number of multiply-adds.
	
	\subsection{Comparisons with state-of-the-art Attention Modules}
	We select a series of state-of-the-art attention modules, which is considered to be relatively lightweight, and demonstrate their performance based on ResNet50 and ResNet101 \cite{he2016deep,he2016identity}. They contain SE \cite{hu2018squeeze}, SK \cite{li2019selective}, BAM \cite{park2018bam}, CBAM \cite{woo2018cbam}, and GC \cite{cao2019GCNet}. For a fair comparison, we implement all the attention modules (partially refer to the official codes\footnote{https://github.com/Jongchan/attention-module, https://github.com/xvjiarui/GCNet}) with their respective best settings using a unified pytorch framework. Following \cite{hu2018squeeze,woo2018cbam}, these attention modules are placed after the last BatchNorm \cite{ioffe2015batch} layer inside each bottleneck except for BAM and SK. BAM \cite{park2018bam} is naturally designed between stages. SK \cite{li2019selective} is originally designed on ResNeXt-like bottlenecks with multiple large-kernel group convolutions. To transfer it to the ResNet backbones, we make a slight modification and only append one additional 3 $\times$ 3 group ($G$ = 32) convolution to each original 3 $\times$ 3 convolution of ResNet, to prevent the parameters and calculations of the corresponding SKNets from being too large or too small. From the results of Table \ref{table_imagenet_cls}, we observe that based on ResNet50, SGE is on par with the best entries from CBAM (Top-1) and SK/SE (Top-5) but has much fewer parameters and slightly less calculations. As for ResNet101, it outperforms most other competing modules with a non-negligible margin. Please note that in our experiments, we find that the GC \cite{cao2019GCNet} module is difficult to train from the beginning, and it will be stuck in a higher loss for a long time before the training loss begins to decline normally. Therefore it does not eventually lead to a high accuracy. In the original paper of GCNet, the authors do not adopt the commonly used training from scratch settings, but finetune the GC module on the well pretrained ResNets to report the results.

	\begin{table}
		\small
		\centering
		\renewcommand\arraystretch{1.2}
		\newcommand{\tabincell}[2]{\begin{tabular}{@{}#1@{}}#2\end{tabular}}
		\caption{Comparisons to the state-of-the-art attention modules on ImageNet validation set. Single 224 $\times$ 224 central crop is adopted for evaluation. All results are reproduced in the pytorch framework. $^{*}$ denotes the modified versions based on ResNet backbones. The best and the second best records are marked as {\bf bold} and {\color{blue}\bf blue}, respectively.}
		\begin{tabular}{l|c|c|c|c}
			\hline
			Backbone&  Param. & GFLOPs & Top-1 Acc (\%) & Top-5 Acc (\%) \\
			\hline
			ResNet50 \cite{he2016deep} & 25.56M & 4.122 & 76.3840 & 92.9080 \\
			\hline
			SE-ResNet50 \cite{hu2018squeeze} & 28.09M & \bf{\color{blue}4.130} & 77.1840 & \bf{\color{blue}  93.6720}\\
			SK-ResNet50$^{*}$ \cite{li2019selective} & 26.15M & 4.185 & 77.5380 & \bf 93.7000\\
			BAM-ResNet50 \cite{park2018bam} & \bf{\color{blue}25.92M} & 4.205 & 76.8980 & 93.4020\\ 
			CBAM-ResNet50 \cite{woo2018cbam} & 28.09M & 4.139 & \bf 77.6260 & 93.6600 \\ 
			GC-ResNet50 \cite{cao2019GCNet} & 28.11M & 4.130 & 73.8880 & 91.6800 \\ 
			SGE-ResNet50  &\bf 25.56M & \bf 4.127 & \bf{\color{blue} 77.5840} & 93.6640\\ 
			\hline\hline
			ResNet101 \cite{he2016deep} & 44.55M & 7.849 & 78.2000 & 93.9060\\ 
			\hline
			SE-ResNet101 \cite{hu2018squeeze} & 49.33M & \bf{\color{blue}7.863} & 78.4680 & 94.1020 \\  
			SK-ResNet101$^{*}$ \cite{li2019selective} & 45.68M & 7.978 & \bf{\color{blue}78.7920} & \bf{\color{blue}94.2680} \\
			BAM-ResNet101 \cite{park2018bam} & \bf{\color{blue}44.91M} & 7.933 & 78.2180 & 94.0180\\ 
			CBAM-ResNet101 \cite{woo2018cbam}  & 49.33M & 7.879 & 78.3540 & 94.0640 \\ 
			GC-ResNet101 \cite{cao2019GCNet} & 49.36M & 7.863 & 74.6420 & 92.0720 \\ 
			SGE-ResNet101 & \bf 44.55M & \bf 7.858 & \bf 78.7980 & \bf 94.3680\\ 
			\hline
		\end{tabular}
		
		\label{table_imagenet_cls}
		\vspace{-16pt}
	\end{table}

	\subsection{Ablation Study}
	In this section, we report the ablation studies on the ImageNet dataset based on SGE-ResNet50, to thoroughly investigate the components of the SGE modules.

	\begin{figure}	
		\small
		\renewcommand\arraystretch{1.2}
		\newcommand{\tabincell}[2]{\begin{tabular}{@{}#1@{}}#2\end{tabular}}
		\begin{minipage}[b]{0.3\linewidth}
			\vspace{+0pt}
			\includegraphics[width=3.6cm]{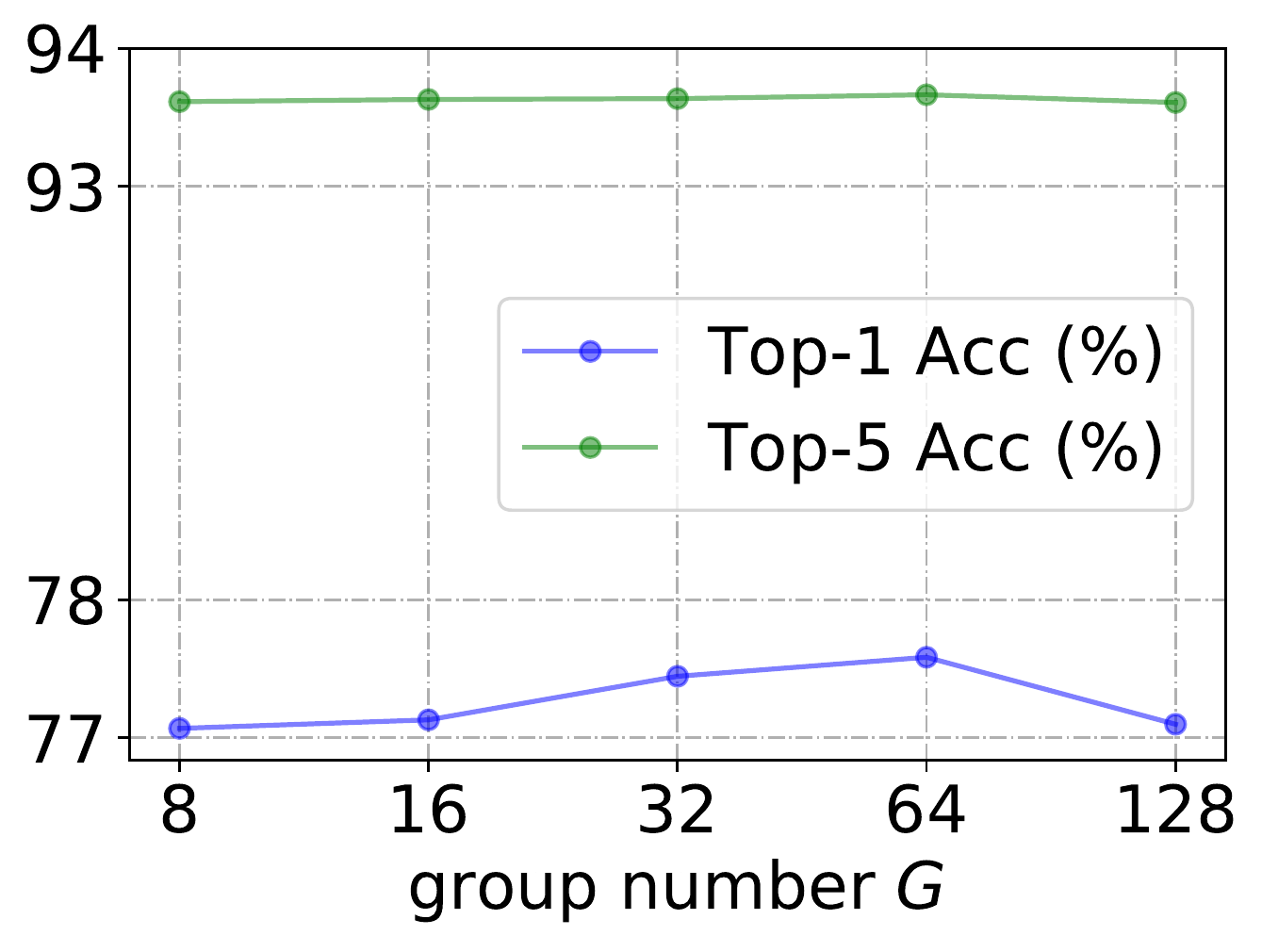}
			\vspace{-6pt}
			\caption{Performance of SGE-ResNet50 as a function of group number $G$.}\label{fig_1a}
		\end{minipage}
		\quad
		\begin{minipage}[b]{0.34\linewidth}
			\centering
			\captionof{table}{Performance of SGE-ResNet50 as a function of initializations of $\gamma$ and $\beta$.}
			\vspace{-7pt}
			\begin{tabular}{l|l|l|l}
				\hline
				\multirow{2}{*}{$\gamma$} & \multicolumn{1}{c|}{\multirow{2}{*}{$\beta$}} & \multicolumn{2}{c}{SGE-ResNet50 Acc}                                             \\ \cline{3-4} 
				& \multicolumn{1}{c|}{}                         & \multicolumn{1}{c|}{Top-1 (\%)} & \multicolumn{1}{c}{Top-5 (\%)} \\ \hline
				0   & 0    & 77.3780 &\bf  93.7140    \\ \hline
				0   & 1    &\bf 77.5840 &  93.6640    \\ \hline
				1   & 0    & 77.2200 &  93.5820    \\ \hline
				1   & 1    & 77.0820 &  93.7040    \\ \hline 
			\end{tabular}
			\vspace{+0pt}
			\label{tab_1b}
		\end{minipage}
		\quad
		\begin{minipage}[b]{0.3\linewidth}

			\captionof{table}{Performance of SGE-ResNet50 with and without the normalization part.}
			\vspace{-7pt}
			\begin{tabular}{c|l|l}
				\hline
				\multirow{2}{*}{Norm} & \multicolumn{2}{c}{SGE-ResNet50 Acc}        \\ \cline{2-3} 
				& \multicolumn{1}{c|}{Top-1 (\%)} & Top-5 (\%) \\ \hline
				w/    &\bf 77.5840 &\bf     93.6640       \\ \hline
				w/o   & 76.4980 &     93.1580       \\ \hline 
			\end{tabular}
			\vspace{+13pt}
			\label{tab_1c}
		\end{minipage}
		\vspace{-20pt}
	\end{figure}

	\textbf{Group number $G$.} In the SGE module, the number of groups $G$ controls the number of different semantic sub-features. Since the total number of channels is fixed, too many groups will result in a reduction in the sub-feature dimension within each group, leading to weaker feature representation for each semantic response; On the contrary, too few groups will make  the diversity of semantics limited. It is natural to speculate that there is a moderate hyperparameter $G$ that balances semantic diversity and the ability of representing each semantic to optimize network performance. From Figure \ref{fig_1a}, we can see that with the increase of $G$, the performance of the network shows a trend of increasing first and then decreasing (especially in terms of Top-1 accuracy), which is highly consistent with our deduction. Through the experimental results, we usually recommend the number of groups $G$ to be 32 or 64. In subsequent experiments, we use $G$ = 64 by default.  
	
	\textbf{Initialization of the $\gamma$ and $\beta$.} During the experiment, we found that the initialization of the parameter $\gamma$ and $\beta$ has a small but not negligible effect on the result. To investigate this, we use values 0, 1 for grid search to see the effects of the initialization. From Table \ref{tab_1b} we find that initializing $\gamma$ to 0 tends to get better results. We speculate that when the ordinary patterns of semantic learning has not yet been completely formulated in convolutional feature maps during the initial stage of network training, it may be appropriate to temporarily discard the attention mechanism, but let the network learn a basic semantic representation first.  After the initial training period, the attention modules then need to be gradually turned in effect. Therefore, in the early moments of network learning, the attention mechanism of SGE is not suggested to participate heavily in training by setting $\gamma$ to 0. Such an operation is almost equivalent to simulate the learning process of a network without attention modules during the very early training stage, since each sub-feature of each location is linearly multiplied by the same constant (i.e., $\mathcal{\sigma}(\beta)$), whose effect can be cancelled by the following BatchNorm layer. 
	
	\textbf{Normalization. } To investigate the importance of normalization in SGE modules, we conduct experiments by eliminating the normalization part from SGE (as shown in Table \ref{tab_1c}) and find that performance is considerably reduced. This confirms our previous conjecture: because the distribution of features generated by different samples for the same semantic group is inconsistent, it is difficult to learn robust importance coefficients without normalization. This is also partially validated in Figure \ref{fig_vd}, where the variance statistic usually has a relatively large standard deviation. It demonstrates that the variance of the activation values of different samples in the same group can be statistically very different, indicating that normalization is essential for SGE to work.
	
	\section{Experiments on Object Detection}
	We further evaluate the SGE module on object detection on COCO 2017 \cite{lin2014microsoft}, whose train set is comprised of 118k images, validation set of 5k images. We follow the standard setting \cite{he2017mask} of evaluating object detection via the standard mean Average-Precision (AP) scores at different box IoUs or object scales, respectively. 
	
	The input images are resized with their shorter side being 800 pixels \cite{lin2017feature}. We train on 8 GPUs with 2 images per each. The backbones of all models are pretrained on ImageNet \cite{russakovsky2015imagenet} (directly borrowed from the models listed in Table \ref{table_imagenet_cls}), then all layers except for the first two stages are jointly finetuned with FPN \cite{lin2017feature} neck and a set of detector heads. Following the conventional finetuning setting \cite{he2017mask}, the BatchNorm layers are frozen during finetuning. All models are trained for 24 epochs using synchronized SGD with a weight decay of 0.0001 and momentum of 0.9. The learning rate is initialized to 0.02, and decays by a factor of 10 at the 18th and 22nd epochs. The choice of hyper-parameters follows the latest release of the detection benchmark \cite{mmdetection2018}.
	
	
	\subsection{Experiments on state-of-the-art Detectors}
	We embed the SGE modules into the popular detector framework separately to check if the enhanced feature map helps to detect objects. We select three popular two-stage detection frameworks, including Faster RCNN \cite{ren2015faster}, Mask RCNN \cite{he2017mask}, and Cascade RCNN \cite{cai2018cascade}, and choose the widely used FPN \cite{lin2017feature} as the detection neck. For a fair comparison, we only replace the pretrained backbone model on ImageNet while keeping the other components in the entire detector intact. Table \ref{tab_compare_detectors} shows the performance of embedding the backbone with the SGE module on these state-of-the-art detectors. We find that although SGE introduces almost no additional parameters and calculations, the gain of detection performance is still very noticeable with basically more than 1\% AP point. It is worth noting that SGE can be more prominently advanced on stronger detectors (\textbf{+1.5\%} AP on ResNet50 and \textbf{+1.8\%} on ResNet101 in Cascade RCNN).
	\begin{table}
		\small
		\centering
		\renewcommand\arraystretch{1.2}
		\newcommand{\tabincell}[2]{\begin{tabular}{@{}#1@{}}#2\end{tabular}}
		\caption{Comparisons based on the state-of-the-art detectors. The Parm. and GFLOPs are only with the backbone parts, given that all the remaining structures are kept the same for a specific detector. The numbers in brackets denote the improvements over the baseline backbones. The SGE modules tend to obtain a larger gain on the stronger baseline detection models. }
		\begin{tabular}{l|c|c|c|l|c|c}
			\hline
			Backbone&  Param. & GFLOPs & Detector & ${\rm AP}_{50:95}$ (\%) & ${\rm AP}_{50}$ (\%) & ${\rm AP}_{75}$ (\%) \\
			\hline
			ResNet50 \cite{he2016deep} & 23.51M & 88.032 & Faster RCNN \cite{ren2015faster} & 37.5 & 59.1 & 40.6\\ 
			SGE-ResNet50 & 23.51M & 88.149
			& Faster RCNN \cite{ren2015faster} & \bf 38.7 {\scriptsize(+1.2)}& \bf 60.8 & \bf 41.7\\
			\hline
			ResNet50 \cite{he2016deep} & 23.51M & 88.032 &Mask RCNN \cite{he2017mask} & 38.6 & 60.0 & 41.9\\ 
			SGE-ResNet50 & 23.51M & 88.149 &Mask RCNN \cite{he2017mask} & \bf 39.6 {\scriptsize(+1.0)} & \bf 61.5 & \bf 42.9\\ 
			\hline
			ResNet50 \cite{he2016deep} & 23.51M & 88.032 &Cascade RCNN \cite{cai2018cascade}&41.1&59.3& 44.8\\  
			SGE-ResNet50 & 23.51M & 88.149 &Cascade RCNN \cite{cai2018cascade}& \bf 42.6 {\scriptsize(+1.5)}&  \bf 61.4& \bf 46.2\\ 
			\hline\hline
			ResNet101 \cite{he2016deep} & 42.50M & 167.908 &Faster RCNN \cite{ren2015faster}& 39.4 & 60.7 & 43.0 \\ 
			SGE-ResNet101 & 42.50M & 168.099 &Faster RCNN \cite{ren2015faster}& \bf 41.0 {\scriptsize(+1.6)} & \bf 63.0 & \bf 44.3 \\  
			\hline
			ResNet101 \cite{he2016deep} & 42.50M & 167.908 &Mask RCNN \cite{he2017mask}& 40.4 & 61.6 & 44.2 \\ 
			SGE-ResNet101 & 42.50M & 168.099 &Mask RCNN \cite{he2017mask}& \bf 42.1 {\scriptsize(+1.7)} & \bf 63.7 & \bf 46.1\\ 
			\hline
			ResNet101 \cite{he2016deep} & 42.50M & 167.908 &Cascade RCNN \cite{cai2018cascade} & 42.6 & 60.9 & 46.4 \\ 
			SGE-ResNet101 & 42.50M & 168.099 &Cascade RCNN \cite{cai2018cascade} & \bf 44.4 {\scriptsize(+1.8)} & \bf 63.2 & \bf 48.4 \\ 
			\hline
		\end{tabular}
		\label{tab_compare_detectors}
		\vspace{-10pt}
	\end{table}

	\subsection{Comparisons with state-of-the-art Attention Modules}
	Next, we chose a representative one-stage detection framework RetinaNet \cite{lin2017focal}, to compare SGE with several competitive state-of-the-art attention modules, especially for objects with three different scales. The original backbones are replaced with the corresponding attention embedded ResNets, which are pretrained on ImageNet, for a fair comparison. In Table \ref{tab_compare_DWC}, SGE greatly improves the accuracy of detection for small objects while its performance of the media and large objects is close to the optimal ones (41.2 \emph{vs} 41.3 from SE and 49.9 \emph{vs} 50.4 from SK), indicating that the SGE module is able to retain the feature representation of the precise spatial area well and is very robust to various object scales. Conversely, the SE/SK module has only a small increase in the recognition of small objects. For SE/SK, in each channel, the same importance coefficient is allocated to each location of the space, probably resulting in the loss of the ability to express the details of micro-regions.
	
	\begin{table}
		\small
		\centering
		\renewcommand\arraystretch{1.2}
		\newcommand{\tabincell}[2]{\begin{tabular}{@{}#1@{}}#2\end{tabular}}
		\caption{Performance on RetinaNet for objects of three scales. The notations are similar as in Table \ref{tab_compare_detectors}. The best and the second best records are marked as {\bf bold} and {\bf\color{blue}blue}, respectively. Compared to the SE/SK module, the detection of small objects from SGE has been significantly improved.}
		\begin{tabular}{l|c|c|l|l|l}
			\hline
			Backbone & Param. & GFLOPs& ${\rm AP}_{\rm small}$ (\%) & ${\rm AP}_{\rm media}$ (\%) & ${\rm AP}_{\rm large}$ (\%)\\
			\hline
			ResNet50 \cite{he2016deep} & 23.51M & 88.032 &19.9&39.6&48.3\\ 
			\hline
			SE-ResNet50 \cite{hu2018squeeze} &  26.04M & {\bf\color{blue}88.152} &20.7 {\scriptsize(+0.8)}&\bf {41.3 {\scriptsize(+1.7)}}&{\bf\color{blue} 50.0 {\scriptsize(+1.7)}}\\ 
			SK-ResNet50 \cite{li2019selective} & 24.11M & 89.414 & 20.2 {\scriptsize(+0.3)}&40.9 {\scriptsize(+1.3)}&\bf {50.4 {\scriptsize(+2.1)}}\\
			BAM-ResNet50 \cite{park2018bam} & {\bf\color{blue} 23.87M} & 89.804 & 19.6 {\scriptsize(-0.3)} & 40.1 {\scriptsize(+0.5)} & 49.9 {\scriptsize(+1.6)}\\
			CBAM-ResNet50 \cite{woo2018cbam} & 26.04M & 88.302 &\bf {21.8 {\scriptsize(+1.9)}} & 40.8 {\scriptsize(+1.2)} & 49.5 {\scriptsize(+1.2)} \\ 
			SGE-ResNet50 &\bf 23.51M &\bf 88.149 &\bf {21.8 {\scriptsize(+1.9)}}& {\bf\color{blue}41.2 {\scriptsize(+1.6)}}&{49.9 {\scriptsize(+1.6)}}\\ 
			\hline
		\end{tabular}
		\label{tab_compare_DWC}
		\vspace{-12pt}
	\end{table}

	\section{Conclusion}
	We propose a Spatial Group-wise Enhance (SGE) module that enables each of its feature groups to autonomously enhance its learnt semantic representation and suppress possible noise, nearly without introducing additional parameters and computational complexity. We visually show that the feature groups have the ability to express different semantics, while the SGE module can significantly enhance this ability. Despite its simplicity, SGE has achieved a steady improvement in both image classification and detection tasks, which demonstrates its compelling effectiveness in practice.

	\small
	\bibliography{neurips_2019}
	\bibliographystyle{plain}

\end{document}